\documentclass{article}

\usepackage[utf8]{inputenc}
\usepackage{mathrsfs}
\usepackage{longtable}
\usepackage{graphicx}
\usepackage{booktabs}
\usepackage{float}
\usepackage{amsmath}
\usepackage{amssymb}

\usepackage[numbers]{natbib}
\title{Evolution and trade-off dynamics of functional load}
\author{Erich Round, Rikker Dockum and Robin J. Ryder}
\date{December 2021}

\begin{document}

\maketitle

\abstract{Function Load (FL) quantifies the contributions by phonological contrasts to distinctions made across the lexicon. Previous research has linked particularly low values of FL to sound change. Here we broaden the scope of enquiry into FL, to its evolution at all values. We apply phylogenetic methods to examine the diachronic evolution of FL across 90 languages of the Pama-Nyungan (PN) family of Australia. We find a high degree of phylogenetic signal in FL. Though phylogenetic signal has been reported for phonological structures, such as phonotactics, its detection in measures of phonological function is novel. We also find a significant, negative correlation between the FL of vowel length and of the following consonant, that is, a deep-time historical trade-off dynamic, which we relate to known  allophony in modern PN langauges and compensatory sound changes in their past. The finding reveals a historical dynamic, similar to transphonologization, which we characterize as a \textit{flow} of contrastiveness between subsystems of the phonology. Recurring across a language family which spans a whole continent and many millennia of time depth, our finding provides one of the most compelling examples yet of Sapir's 'drift' hypothesis, of non-accidentally parallel development in historically related languages.}


\section{Introduction}

Function Load (FL) quantifies the contribution of specific phonological contrasts to distinctions made in the lexicon of a language \citep{martinet_function_1952,hockett_quantification_1967,surendran_quantifying_2006}. In English for example, the phonemes /t/ and /d/ contrast, and thus there exist phonological strings in English, including whole words, which differ only by virtue of one containing /t/ in a position where the other contains /d/. Examples include \textit{time/dime}, \textit{welter/welder} and \textit{hit/hid}. At a conceptual level, the FL of the \{/t/,/d/\} contrast in English is the degree to which distinctions in the English lexicon are enabled by that contrast, or conversely, the degree to which they would be conflated if /t/ and /d/ were merged into a single category.

A classic operationalization of FL by \citet{hockett_quantification_1967} is in terms of entropy \citep{shannon_mathematical_1948}. Hockett's definition makes reference to domains, $\mathcal{D}$, which could be words, morphemes, syllables, or any kind of substring composed of phonemes. In a language $\mathscr{L}$, the lexicon $\Lambda$ will contain a set $\mathbf{S}_{\mathcal{D},\Lambda}$ of unique phonological string types, $s$, which comprise a domain of type $\mathcal{D}$. The entropy of domain $\mathcal{D}$ in lexicon $\Lambda$ is:

\begin{equation}
H_{\mathcal{D},\Lambda} = - \sum_{s \in \mathbf{S}_{\mathcal{D},\Lambda}} \text{P}(s).\log_2\text{P}(s)
\end{equation}

The FL of a phonological contrast $\phi$ in lexicon $\Lambda$ and domain $\mathcal{D}$ is the difference between two entropy measures: the entropy of domain $\mathcal{D}$ in lexicon $\Lambda$, and of domain $\mathcal{D}$ in an altered lexicon $\Lambda^\prime_\phi$, created by collapsing the contrast $\phi$ in $\Lambda$:

\begin{equation}
f(D,\Lambda,\phi) = H_{\mathcal{D},\Lambda} - H_{\mathcal{D},\Lambda^\prime_\phi}
\end{equation}

A phonological contrast, $\phi$, may refer to distinctions between the members of a single set of phonemes, such as between two phonemes \{/t/, /d/\} or between four phonemes \{/t/, /d/, /s/, /z/\}. Alternatively, it may refer to a collection of multiple, parallel distinctions, defined by a collection of sets and the distinctions within each of them. For instance, a contrast between voiced and voiceless stop consonants could refer to the distinctions within each of the three sets in the collection \{\{/p/, /b/\}, \{/t/, /d/\}, \{/k/, /g/\}\} and a contrast between the places of articulation of stop consonants could refer to the distinctions within each of the two sets in the collection \{\{/p/, /t/, /k/\}, \{/b/, /d/, /g/\}\}. In all cases, the altered lexicon $\Lambda^\prime_\phi$ is obtained by taking each individual set and replacing the phonemes within it with a single phonemic symbol which is distinct from all other phonemic symbols in $\mathscr{L}$.

\subsection{Functional Load and sound change}

Languages undergo mutational changes, known as sound change, in which sounds may change from one phonemic category to another \citep{garrett_sound_2015}. The term \textit{unconditioned merger} refers to sound changes which cause two or more previously distinct phonemic categories to become conflated into one. Conditioned mergers are when conflation affects the phonemes only in certain contexts. FL has attracted attention as a possible explanatory factor in the incidence of mergers, with recent evidence supporting a much older conjecture that contrasts with low FL are more prone to merge than contrasts with high FL \citep{martinet_function_1952,silverman_neutralization_2010,bouchard-cote_automated_2013,wedel_high_2013,babinski_mergers_2018}. Debate is ongoing over which operationalizations of FL provide the greatest predictive power \citep{wedel_high_2013} and what kinds of mergers FL predicts \citep{ceolin_functional_2020}. Whether it has used entropy-based definitions of FL or not, most research to date has focused on domains $\mathcal{D}$ that are words.

When a contrast undergoes unconditional merger, its FL falls to zero. The fact that all else equal, contrasts with low FL values are more likely to fall to zero than contrasts with higher FL values, can be viewed as an expected outcome of any system in which FL evolves over time according to a stochastic process in which small changes are more likely than large ones. Such a system leads to other predictions too. A novel contribution of this study is to test such predictions, and thus to broaden the scope of research into FL. 

Here we examine FL from a phylogenetic perspective, investigating how FL evolves over time in the large, Pama-Nyungan (PN) language family of Australia. PN languages extend across 90\% of the Australian mainland and the time depth of the family is estimated at around 5--6,000 years Before Present \citep{mcconvell_backtracking_1996,bowern_computational_2012,bouckaert_origin_2018}. In Study 1, we apply standard techniques to show that FL contains significant phylogenetic signal, which is to say, the variance in the FL data across the PN language family is consistent with the hypothesis that FL has evolved along the PN tree according to stochastic process that is well approximated by a Brownian motion.

\subsection{Contrastiveness in Pama-Nyungan VC strings}

In many PN languages, an inverse correlation has been observed between phonemic vowel length and the phonetic duration of a following consonant \citep{butcher_what_1999,tabain_vc_2004,butcher_australian_2006,fletcher_sound_2014,jepson_vowel_2015}. This is particularly so for vowels in the first syllable of words, referred to as tonic vowels. Here we focus on tonic vowels and single, intervocalic consonants which follow them. Post-tonic single consonants which follow phonemically short vowels and which have phonetically longer durations in some language also exhibit additional phonetic properties associated with long duration, such as more complete closure and passive devoicing of stops, and pre-stopping of nasals and laterals. Conversely, post-tonic single consonants which follow phonemically long vowels and which have phonetically shorter durations may exhibit more voicing and lenition of stops, and the absence of pre-stopping in laterals and nasals \citep{hercus_pre-stopped_1972,butcher_what_1999,tabain_vc_2004,loakes_phonetically_2008,round_prestopping_2014}. Examples of allophonic conditioning of this kind are cited in Table \ref{PN-allophony}.

\begin{table}[H] 
\caption{Examples of allophony in post-tonic consonants, conditioned by phonemic length of the tonic vowel. Key: V\_ after phonemically short vowel, VV\_ after phonemically long vowel. See references for additional details in the conditioning of allophony.\label{PN-allophony}}
\begin{tabular}{llll}
\toprule
\textbf{Language (subgroup)} & \textbf{Consonants}	& \textbf{V\_} & \textbf{VV\_}\\
\midrule
Djambarrpuyngu (Yolngu) \citep{jepson_vowel_2015} & Consonants & longer & shorter\\
Wik (Middle Paman) \citep{hale_wik_1976} & Stops & tenser & laxer\\
Kugu Nganhcara (Middle Paman) \citep{smith_kugu_2000} & Voiced stops & stop & fricative\\
Nukunu (Thura-Yura) \citep{hercus_nukunu_1992} & Nasals, Laterals & prestopped & plain \\ 
Yadhaykenu (Nothern Paman) \citep{crowley_uradhi_1983} & Laterals & plain & flapped\\
\bottomrule
\end{tabular}
\end{table}

A general fact about sound change is that when two phonemes merge, it is possible for phonetic correlates of the original contrast, which are manifested in other segments, to remain in place and become contrastive. This has occurred in multiple branches of PN, as phonemic vowel length is lost while its erstwhile phonetic correlates on the following consonant remain and become distinctive. Examples are cited in Table \ref{PN-changes}.

\begin{table}[H] 
\caption{Examples of post-tonic consonant contrasts created upon the loss of length distinctions in tonic vowels of Pama-Nyungan languages. Key: T short stop, TT long stop, D voiced stop, Z spirant, N nasal, NN long nasal, DN prestopped nasal, ND nasal+stop, L lateral, DL prestopped lateral, V\_ after erstwhile short vowel, VV\_ after erstwhile long vowel. See references for additional details and conditioning of the tabulated sound changes.\label{PN-changes}}
\begin{tabular}{llll}
\toprule
\textbf{Language (subgroup)} & \textbf{Original C}	& \textbf{V\_} & \textbf{VV\_}\\
\midrule
Warumungu (Warunmungic) \citep{alpher_sound_2022} & T & TT & T\\
Wik-Muminh (Middle Paman) \citep{hale_wik_1976} & T (non-apical) & T & D\\
Northern Paman subgroup \citep{hale_phonological_1976,hale_uradhi_1976,crowley_uradhi_1983} & T (non-apical) & T & Z\\
Lamalama, Umbuygamu (Lamalamic) \citep{verstraete_lamalamic_2018} & /k/ & /k/ & /h/\\
Kugu Mumminh (Middle Paman) \citep{alpher_sound_2022} & N & NN & N \\ 
Arandic subgroup \citep{koch_pama-nyungan_1997,koch_basic_2001} & N & DN & N\\
Walangama (Norman Paman) \citep{black_norman_1980} & N & DN & N\\
Olgolo (Southwest Paman) \citep{dixon_olgolo_1970} & N & DN & N\\
Lamalama (Lamalamic) \citep{verstraete_lamalamic_2018} & N & ND & N\\
Rimanggudinhma (Lamalamic) \citep{verstraete_lamalamic_2018} & N & D & N\\
Thura-Yura subgroup \citep{hercus_nukunu_1992} & L & DL & L\\
\bottomrule
\end{tabular}
\end{table}

In the cases cited in Table \ref{PN-changes}, the complete loss of vowel length is associated with an increase in contrastiveness in following consonants. In such cases, the FL of the vowel length contrast falls (to zero) while the FL of the following consonant position rises. Consequently, there is a trade-off relationship between between $FL_V$, the FL of vowel length, and $FL_C$, the FL of the manner of articulation (included voicing and fortition) of the following consonant. This trade-off is of a very specific kind, in which the complete merger of all short/long vowel pairs reduces $FL_V$ to zero. We will refer to this as a \textit{trade-off with contrast collapse}. Other trade-offs are possible, however. If a length contrast is lost only in certain vowels, and/or only in certain contexts, then $FL_V$ would fall (though not to zero) and in such cases, $FL_C$ could be expected to rise if consonants become more contrastive when they follow the vowels which do merge. This scenario would give rise to a second kind of trade-off. We will refer to this second kind as a \textit{trade-off with contrast maintenance}. 

In Study 2, we test the hypothesis that $FL_V$ and $FL_C$ are negatively correlated in PN. Recall that $FL_C$ relates to post-tonic consonants and is defined in terms of their manner of articulation. We expect $FL_C$ to correlate negatively with $FL_V$ for the phonetic and historical reasons introduced above. We also examine $FL_P$, the FL of post-tonic consonants' place of articulation. Because research on PN languages has identified no particular association between vowel length and consonant place, our hypothesis is that $FL_P$ will show no significant correlation with $FL_V$.

\section{Materials and Methods}

\subsection{Functional load data in Pama-Nyungan}

\begin{figure}
\caption{Pama-Nyungan tree containing the 90 languages used in this study. Displayed here is a single, \textit{maximum clade credibility} tree, i.e., the one tree within the 1000-tree sample which most adequately represents the highest-probability subgroups in all of the trees of the sample.\label{tree}}
\includegraphics[width=6.5cm, angle=270]{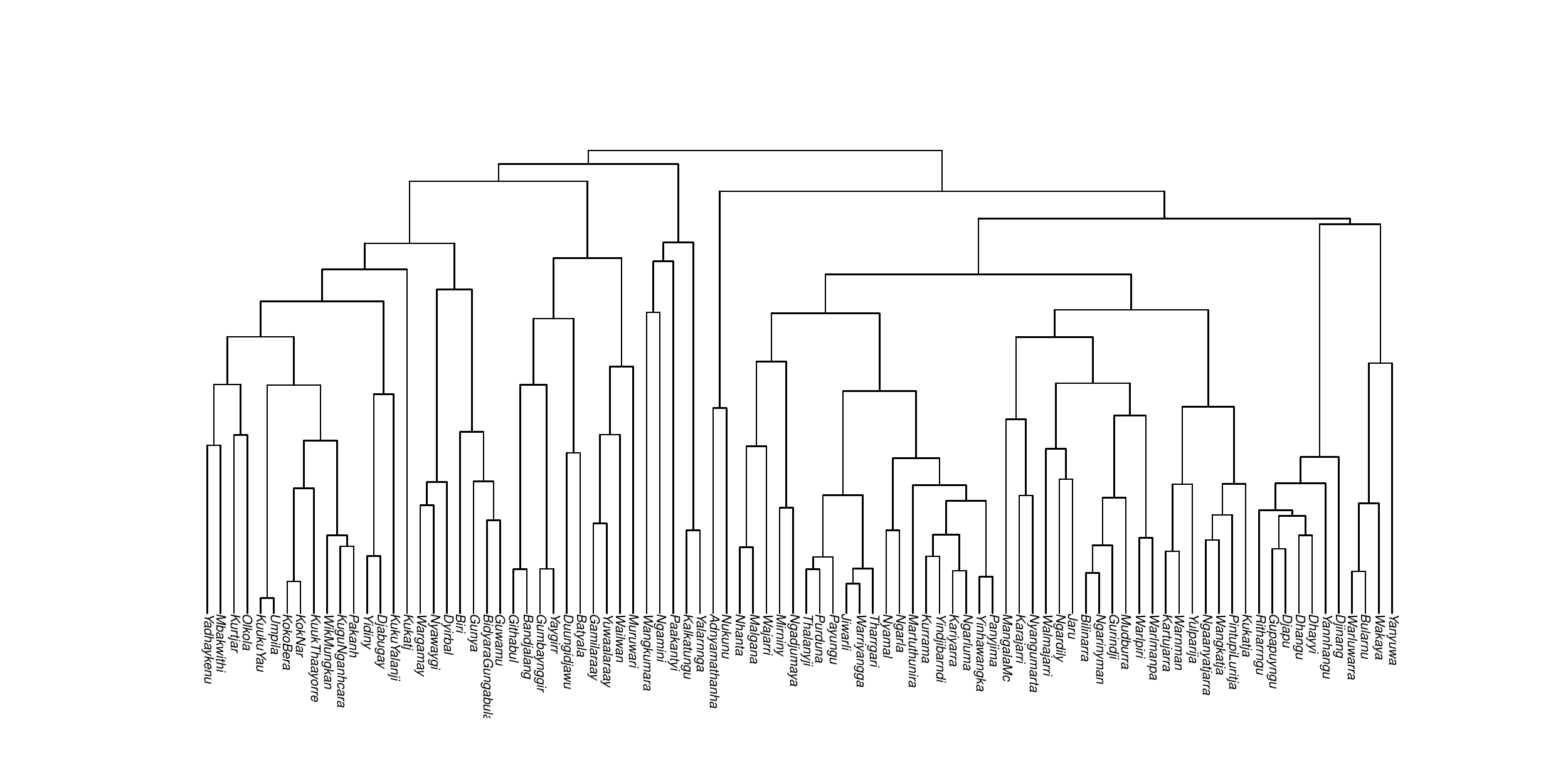}
\end{figure}

FL was estimated for vowel length ($FL_V$), consonant manner ($FL_C$) and consonant place ($FL_P$) in domains comprised of a tonic vowel followed by a single, intervocalic consonant, in a set of $90$ Pama-Nyungan languages listed in Appendix A. FL estimates were based on lexical datasets, which contained between 208 and 3215 tokens of the domains of interest (mean 774, median 605). The $90$ languages studied were selected by taking the $112$ PN lexicons studied in \citep{macklin-cordes_phylogenetic_2021} and removing all whose $FL_V$ was zero or which had fewer than 200 domain tokens. A tree of these $90$ languages is shown in Figure \ref{tree}.

When classifying vowels as short or long, we regarded sequences of two adjacent short vowels as one long vowel, and sequences of /uwu/ and /iji/ as long vowels, since a tradition followed in some Australianist analysis is to represent long high vowels [u:] and [i:] as phonemic vowel-glide-vowel sequences (e.g. \citep[p.24]{austin_grammar_1981}, \citep[p.91]{mcgregor_functional_1990}).

Phonemically long or geminate consonants, and phonemically pre-stopped sonorants, have been analysed as both mono- and bi-segmental units in the Australianist literature \citep{round_phonotactics_2022}. Here, we were guided by the kinds of historical developments that we wish to study, and classed them as single segments.

The FL data obtained for the $90$ PN languages is reported in Appendix A.

\subsection{Phylogenetic analysis}

Because languages are related to one another, it is not statistically valid to treat cross-linguistic observations as independent \citep{macklin-cordes_challenges_nodate}. Quantitative phylogenetic methods \citep{garamszegi_modern_2014} take phylogenetic relatedness into account in a principled and statistically sound manner. Our two studies use phylogenetic techniques in order to make valid inferences from the cross-linguistic FL data in PN.

Study 1 assesses the degree of phylogenetic signal in $FL_V$, $FL_C$ and $FL_P$. Phylogenetic signal is a measure of genealogical structuration in the variance in the data. Here we measure phylogenetic signal using a widely employed statistic, Blomberg's $K$ \citep{blomberg_testing_2003,macklin-cordes_phylogenetic_2021}. Intuitively, Blomberg's $K$ measures how well the observed distribution of FL values accords with the hypothesis that they have evolved along the PN family tree according to a stochastic, Brownian motion process. 

To calculate phylogenetic signal, reference must be made to a PN family tree, however in linguistics there is uncertainty about the details of this tree. Here we employ a standard approach to accounting for phylogenetic uncertainty of this kind, by measuring phylogenetic signal with respect to not one tree, but a sample of 1000 highly-likely family trees \citep{bowern_pama-nyungan_2015}. This generates 1000 estimates of $K$, providing a distribution describing its likely value. Blomberg's $K$ takes a value of $1$ if the data accord perfectly with the reference tree and a minimum value of $0$ if the data is perfectly randomly distributed. Values in excess of $1$ are possible if FL data values are highly clumped within subgroups of the family.

Study 2 examines the phylogenetic Pearson's correlation \citep{martins_phylogenetic_1991} between $FL_V$ and $FL_C$ and between $FL_V$ and $FL_P$. This test is conceptually parallel to a regular Pearson’s correlation, but also takes into account the specific kinds of non-independence caused by genealogical relationships between the languages. As with our estimate of phylogenetic signal, we take into account phylogenetic uncertainty by performing the correlation test in reference to the sample of 1000 highly-likely trees.

We used 1000 dated phylogenetic trees from the posterior distribution provided by \citet{bowern_pama-nyungan_2015}; these trees were reconstructed using cognate data. We assume that $(FL_V, FL_C, FL_p)$ evolves along a phylogeny following a Brownian motion. We used the \texttt{phytools} R package \citep{revell2012phytools} to estimate the covariance matrix of the Brownian motion on each tree in the sample, giving a sample from the posterior distribution of Pearson's $r$ correlation. The $p$-values reported were computed using the posterior mean estimates.

\section{Results}

Phylogenetic signal as measured by Blomberg’s $K$ was very close to $1$ for $FL_V$, $FL_C$ and $FL_P$, as shown in Table \ref{phylo-sig-results}. To place these $K$ values in context, \citet{macklin-cordes_phylogenetic_2021} examine the lexical Markov chain transition probabilities of biphones (two-segment sequences) in PN and find mean $K$ values of $0.54$, or mean $K$ of $0.59$ when segments were binned into groups by place or manner of articulation; \citet{macklin-cordes_challenges_nodate} examine relative frequencies of dental versus palatal consonants in word initial and intervocalic positions in PN and find mean $K$ values from $0.78$ to $1.32$ word-initially and from $0.34$ to $0.70$ intervocalically; \citet{dockum_phylogeny_2018} examines phoneme frequencies and biphone Markov chain transition probabilities in languages of the Tai family and finds mean $K$ of $0.71$ and $0.68$ respectively. Further afield, \citet{blomberg_testing_2003} examined 121 biological traits of a wide variety of plant and animal organisms, finding mean $K$ of $0.35$ for behavioral traits, $0.54$ for physiology and $0.83$ for traits related to body size. Taken in this context, our results suggest that the evolution of FL is very well modelled by a stochastic, Brownian motion process along the PN tree.

\begin{table}[H] 
\caption{Phylogenetic signal in $FL_V$, $FL_C$ and $FL_P$, measured using Blomberg's $K$ and a sample of 1000 reference PN trees.\label{phylo-sig-results}}
\begin{tabular}{lll}
\toprule
\textbf{Functional load measure} & \textbf{mean $K$}	& \textbf{std.dev of $K$}\\
\midrule
FL of tonic vowel length ($FL_V$) & $0.972$ & $0.036$ \\
FL of post-tonic consonant manner ($FL_C$) & $0.956$ & $0.038$ \\
FL of post-tonic consonant place ($FL_P$) & $0.960$ & $0.030$ \\
\bottomrule
\end{tabular}
\end{table}

One consequence of the high levels of phylogenetic signal found in FL, is that statistical analysis, such as the measurement of correlations, should be carried out using phylogenetic comparative methods \citep{macklin-cordes_challenges_nodate}. Phylogenetic Pearson’s correlation (Table \ref{phylo-cor-results}) was significant and negative between $FL_V$ and $FL_C$, but did not reach significance between $FL_V$ and $FL_P$, in accordance with our hypotheses.

\begin{table}[H] 
\caption{Phylogenetic Pearson’s correlation between $FL_V$ and $FL_C$ and $FL_P$.\label{phylo-cor-results}}
\begin{tabular}{llcl}
\toprule
\textbf{Functional load measures} & \textbf{$r$}	& 95\% interval & \textbf{$p$}\\
\midrule
$FL_V$ versus $FL_C$ & $-0.28$ & $[-0.31\ -0.26]$ & $0.006$ \\
$FL_V$ versus $FL_P$ & $0.03$ & $[0.001\ 0.055]$ &  $0.778$ \\
\bottomrule
\end{tabular}
\end{table}

\section{Discussion}

Our study examines language diachrony at a statistical level. By doing so, it contributes to a more precise, quantitative characterization of diachronic typology. Specifically, here we have studied the historical dynamics of FL, which is a quantitative characterization of the contribution of specific contrasts to distinctiveness in the lexicon. We have established the potential for these contributions to be correlated with one another, evolving in a statistically non-independent fashion across almost a hundred languages and thousands of years of history. Moreover, we have demonstrated that all values of FL, not only those close to zero, exhibit interesting historical dynamics deserving of more investigation. In this discussion section we select three points that warrant additional emphasis.

\subsection{High degree of phylogenetic signal in FL}
Phylogenetic signal has recently been shown to be present in phonotactic biphone frequencies, phoneme frequencies, and contextual ratios of places of articulation \citep{macklin-cordes_high-definition_2015,dockum_phylogeny_2018, dockum2019tonal,macklin-cordes_phylogenetic_2021,macklin-cordes_challenges_nodate}. These studies reveal that the frequencies of phonological \textit{structures} pattern with genealogy. Here we find a high level of phylogenetic signal also in FL, that is, in the contrastive \textit{function} that phonological structures serve. Interestingly, we find that phylogenetic signal in the FL measures examined here is very close to $1$, and closer than the values found in studies of phonological structures. Why this is so, and how far the finding generalizes, is not yet apparent to us, and deserves further study.

\subsection{Transphonologization and the flow of contrastiveness}
Transphonologization \citep{haudricourt_les_1968} (see also cheshirization \citep{matisoff_areal_1991}) is a term given to sound changes in which a contrastive function is preserved, but the locus of the contrast -- the segments or features which instantiate it -- changes. Here we have studied a closely related phenomenon, in which contrasts do not necessarily disappear or emerge in their entirety, but the relative contrastive workload of them (their FL) does shift from one to another. One way to view this phenomenon is in terms of a diachronic \textit{flow} of FL from one contrast to another (cf \citep{macklin-cordes_re-evaluating_2020}). The fact that we are able to quantitatively detect the presence of this flow of contrastiveness through a language family as large and old as PN suggests the potential of new avenues for investigating the dynamic flow of contrastiveness through phonological systems as they evolve over time.

One question that arises, is whether our findings in PN might reflect some strong preference in language for the conservation of contrastiveness, in which case the flow of FL from one contrast to another might be regarded as an automatic consequence of one contrast undergoing a significant decrease in FL. Although our results alone cannot answer this question, we doubt that such a principle exists in any strong form. Certainly, in many mergers, the overall contrastive capacity of a language is simply reduced, as the FL of one contrast falls but no other FL rises to balance it. In the case of PN tonic vowels and post-tonic consonants, we suggest that the cause of recurrent historical flow of FL lies in particular phonetic factors that are common across PN languages: a synchronic correlation between phonetic tonic vowel duration and phonetic post-tonic consonant manner, even in systems in which only the vowel-durational aspect is tied to a synchronic phonemic contrast; when the phonetic vowel-durational differences are neutralized diachronically, causing the phonemic vowel length contrasts to collapse, the phonetic manner differences -- which still correlate with the same lexical distinctions which vowel length had signalled --  become phonemic. On this view, it is the phonetics of the vowel-consonant strings that furnish the conditions for a natural flow of FL from vowel to consonant.

\subsection{On Sapir's 'drift': the non-accidental, parallel evolution of related languages}
It is a century now since the appearance in print of Edward Sapir’s hypothesis that languages undergo parallel grammatical evolution for several centuries after they split \citep{sapir_language:_1921}. Providing anything more than anecdotal evidence in support of Sapir's hypothesis has long been difficult \citep{grierson_modern_1931,fortescue_drift_2006,croft_explaining_2000}, and some apparent cases may be due to language contact \citep{dahl_growth_2004,heine_language_2005}. \citet{dunn_evolved_2011} used phylogenetic methods to show different language families undergoing different patterns of word order evolution, however the study did not produce an identifiable cause for those patterns. 

Ideally, evidence in support of Sapir's drift should not be anecdotal, but rather be statistically significant across a language family; it should not be reducible to the effects of language contact; and it should be relatable to an identifiable cause. The current study meets these three criteria. It detects parallel changes in FL that are instantiated statistically across $90$ languages, within the PN family whose time depth is estimated at around 5--6,000 years \citep{mcconvell_backtracking_1996,bowern_computational_2012,bouckaert_origin_2018}, thus the evidence is not anecdotal. The data pattern tightly with phylogeny, so are unlikely to be due to contact. And we have identified a causal basis for them, in the common phonetics of PN tonic vowel-consonant sequences. Thus, we believe our results to be one of the fullest confirmations yet, that Sapir's conjecture was essentially correct: that under the right circumstances, languages can undergo parallel grammatical evolution, not merely for centuries but for millennia.

It has been suggested by \citet{joseph2013demystifying} that drift in phonology may be due to a narrowing of the range of variation inherited from a proto-language. In the PN changes described here, however, the flow of FL from vowel length to consonant manner is not due to any narrowing of variation in FL in proto-PN (indeed it is not entirely clear what it should mean for FL to have a range of variation). Nor, when the PN developments are viewed in terms of phonological substance, are they a matter merely of narrowing variation. Although contrasts in vowel length are lost, new variation is introduced in the inventory of contrastive consonant manners and into the set of relationships that can hold between the length of a tonic vowel and the manner of post-tonic consonants. 

In reality, Joseph's proposal would appear to reduce to a fact, well-recognised in evolutionary biology, that incomplete lineage sorting (i.e., the inheritance of variation from a proto-taxon into its descendant) can result in the appearance of convergent evolution \citep{mendes_gene_2016}. But this does not entail that all convergent evolution is due to incomplete lineage sorting (see also \citep{iosad_phonological_2021}). Another important source can be the existence of dependencies with a system which are inherited along with its substance \citep{round_getting_2019}, and which will favour certain outcomes over others in descendent systems, as has been observed in protein evolution for example \citep{storz_causes_2016}. In PN, certain phonetic dependencies between tonic vowel length and post-tonic consonant manners were inherited alongside the phonological substance itself. In the descendent systems, in the event that vowel length was lost, the inherited dependencies favoured the rise of new, contrastive consonant manners.

\section{Conclusions}

This paper joins a growing body of work on the application of computational phylogenetic methods to phonological data. It also represents the first phylogenetic study of FL. We show that there is significant phylogenetic signal in FL, which has implications for a better understanding of the dynamics of sound change. Further, we show that the FL of tonic vowels and and post-vowel consonants are negatively correlated in PN, and that this maps closely to a sample of highly probably PN family trees. We also introduce the idea of the \textit{flow} of contrastiveness between subsystems of the phonology in different languages, which is connected to the concept of transphonologization, and claim that this represents a concrete example of Sapir's drift.

Setting our gaze beyond PN, although not all sound changes are associated with phonetic conditions that promote the flow of FL, there are many do appear to be, and it will be valuable to apply the methods we have introduced here to study them also. In time, this may lead to a more general understanding of how FL can flow within phonological systems over long time horizons. Promising future applications of our approach include the investigation of other suspected diachronic trade-offs such as the rise of phonemic tone and register (such as contrastive phonation) in Southeast Asia, tied to losses of consonantal laryngeal distinctions \citep{matisoff_tonogenesis_1973,huffman_register_1976,dockum_east_2021,surendran_quantifying_2006}.

\section{Appendix A. Functional load data}

\begin{table}[H] 
\caption{Functional load estimates and number of observations on which they are based in Pama-Nyungan languages A--M.}
\begin{tabular}{llllr}
\toprule
Language         & $FL_V$   & $FL_C$   & $FL_P$   & N    \\
\midrule
Adnyamathanha     & 0.007 & 2.299 & 1.693 & 1424 \\
Bandjalang        & 0.388 & 1.211 & 0.846 & 864  \\
Batyala           & 0.153 & 1.205 & 0.840 & 303  \\
Bidyara-Gungabula & 0.020 & 1.044 & 1.439 & 466  \\
Bilinarra         & 0.024 & 1.601 & 1.385 & 1071 \\
Biri              & 0.011 & 1.047 & 1.308 & 387  \\
Bularnu           & 0.010 & 1.814 & 1.532 & 514  \\
Dhangu            & 0.537 & 1.306 & 1.204 & 222  \\
Dhayyi            & 0.620 & 1.335 & 1.356 & 636  \\
Djabugay          & 0.051 & 1.269 & 1.161 & 641  \\
Djapu             & 0.552 & 1.426 & 1.456 & 563  \\
Djinang           & 0.008 & 1.757 & 1.276 & 1459 \\
Duungidjawu       & 0.359 & 0.932 & 0.913 & 340  \\
Dyirbal           & 0.019 & 1.207 & 1.226 & 339  \\
Gamilaraay        & 0.476 & 1.092 & 0.667 & 619  \\
Githabul          & 0.383 & 1.209 & 0.850 & 863  \\
Gumbaynggir       & 0.682 & 1.142 & 0.561 & 260  \\
Gunya             & 0.086 & 1.675 & 1.879 & 421  \\
Gupapuyngu        & 0.703 & 1.660 & 1.390 & 1426 \\
Gurindji          & 0.141 & 1.684 & 1.400 & 2783 \\
Guwamu            & 0.152 & 0.972 & 1.283 & 284  \\
Jaru              & 0.146 & 1.651 & 1.239 & 1459 \\
Jiwarli           & 0.107 & 1.500 & 1.545 & 747  \\
Kalkatungu        & 0.174 & 1.451 & 1.667 & 937  \\
Karajarri         & 0.011 & 1.585 & 1.372 & 1217 \\
Kariyarra         & 0.008 & 1.512 & 1.342 & 252  \\
Kartujarra        & 0.103 & 1.598 & 1.629 & 526  \\
Kok Nar           & 0.024 & 1.025 & 1.038 & 213  \\
Koko Bera         & 0.008 & 1.125 & 1.144 & 428  \\
Kugu Nganhcara    & 0.231 & 1.201 & 1.524 & 359  \\
Kukatj            & 0.492 & 1.371 & 1.262 & 422  \\
Kukatja           & 0.177 & 1.692 & 1.556 & 2339 \\
Kuku Yalanji      & 0.006 & 1.206 & 1.064 & 1070 \\
Kurrama           & 0.157 & 1.475 & 1.333 & 495  \\
Kurtjar           & 0.357 & 1.098 & 0.902 & 405  \\
Kuuk Thaayorre    & 0.398 & 0.946 & 1.297 & 873  \\
Kuuku Ya'u        & 0.822 & 0.968 & 1.827 & 672  \\
Malgana           & 0.091 & 1.360 & 1.520 & 208  \\
Mangala           & 0.068 & 1.585 & 1.312 & 749  \\
Martuthunira      & 0.122 & 1.597 & 1.413 & 633  \\
Mbakwithi         & 0.158 & 0.890 & 0.946 & 255  \\
Mirniny           & 0.057 & 1.494 & 1.646 & 259  \\
Mudburra          & 0.121 & 1.573 & 1.362 & 509  \\
Muruwari          & 0.299 & 1.303 & 1.239 & 873  \\
\bottomrule
\end{tabular}
\end{table}

\begin{table}[H] 
\caption{Functional load estimates and number of observations on which they are based in Pama-Nyungan languages N--Z.}
\begin{tabular}{llllr}
\toprule
Language         & $FL_V$   & $FL_C$   & $FL_P$   & N    \\
\midrule
Ngaanyatjarra     & 0.308 & 1.657 & 1.544 & 1125 \\
Ngadjumaya        & 0.279 & 1.581 & 1.525 & 512  \\
Ngamini           & 0.009 & 1.658 & 1.628 & 453  \\
Ngardily          & 0.044 & 1.541 & 1.419 & 274  \\
Ngarinyman        & 0.050 & 1.609 & 1.371 & 870  \\
Ngarla            & 0.054 & 1.650 & 1.522 & 940  \\
Ngarluma          & 0.009 & 1.550 & 1.511 & 633  \\
Nhanta            & 0.221 & 1.617 & 1.983 & 427  \\
Nukunu            & 0.230 & 1.885 & 1.572 & 282  \\
Nyamal            & 0.011 & 1.668 & 1.585 & 574  \\
Nyangumarta       & 0.056 & 1.637 & 1.530 & 1002 \\
Nyawaygi          & 0.448 & 0.917 & 0.847 & 339  \\
Olkola            & 0.009 & 1.897 & 1.525 & 992  \\
Paakantyi         & 0.482 & 1.310 & 1.586 & 748  \\
Pakanh            & 0.482 & 0.757 & 1.702 & 379  \\
Panyjima          & 0.033 & 1.526 & 1.537 & 327  \\
Payungu           & 0.120 & 1.443 & 1.487 & 514  \\
Pintupi-Luritja   & 0.257 & 1.620 & 1.498 & 3065 \\
Purduna           & 0.186 & 1.484 & 1.641 & 540  \\
Ritharrngu        & 0.528 & 1.594 & 1.285 & 841  \\
Thalanyji         & 0.113 & 1.412 & 1.640 & 467  \\
Tharrgari         & 0.101 & 1.808 & 1.468 & 371  \\
Umpila            & 0.726 & 0.895 & 1.792 & 473  \\
Wailwan           & 0.441 & 1.112 & 0.699 & 489  \\
Wajarri           & 0.096 & 1.555 & 1.585 & 787  \\
Wakaya            & 0.026 & 1.478 & 1.347 & 696  \\
Walmajarri        & 0.075 & 1.695 & 1.531 & 2361 \\
Wangkatja         & 0.243 & 1.647 & 1.549 & 1290 \\
Wangkumara        & 0.037 & 1.784 & 1.492 & 480  \\
Wargamay          & 0.466 & 1.166 & 1.224 & 470  \\
Warlmanpa         & 0.032 & 1.769 & 1.546 & 603  \\
Warlpiri          & 0.078 & 1.717 & 1.549 & 3215 \\
Warluwarra        & 0.062 & 1.625 & 1.609 & 731  \\
Warnman           & 0.034 & 1.619 & 1.557 & 607  \\
Warriyangga       & 0.049 & 1.463 & 1.557 & 273  \\
Wik Mungkan       & 0.579 & 0.911 & 1.733 & 1411 \\
Yadhaykenu        & 0.118 & 1.390 & 1.474 & 385  \\
Yalarnnga         & 0.013 & 1.479 & 1.596 & 397  \\
Yannhangu         & 0.527 & 1.621 & 1.351 & 952  \\
Yanyuwa           & 0.028 & 1.534 & 1.477 & 1351 \\
Yaygirr           & 0.627 & 1.490 & 1.004 & 658  \\
Yidiny            & 0.014 & 1.240 & 1.084 & 964  \\
Yindjibarndi      & 0.127 & 1.476 & 1.212 & 492  \\
Yinhawangka       & 0.080 & 1.654 & 1.678 & 773  \\
Yulparija         & 0.078 & 1.657 & 1.582 & 1234 \\
Yuwaalaraay       & 0.610 & 1.185 & 0.807 & 1109 \\
\bottomrule
\end{tabular}
\end{table}

\bibliography{references}

\end{document}